\pdfoutput=1

\documentclass[11pt]{article}

\usepackage[final]{EACL2023}

\usepackage{times}
\usepackage{latexsym}

\usepackage[T1]{fontenc}

\usepackage[utf8]{inputenc}

\usepackage{microtype}

\usepackage{inconsolata}
\usepackage{algorithm}
\usepackage{algorithmic}

\title{A Group-Specific Approach to NLP for Hate Speech Detection}

\author{Karina Halevy \\
  Harvard University \\
  \texttt{khalevy@college.harvard.edu}}

\begin{document}
\maketitle
\begin{abstract}
Automatic hate speech detection is an important yet complex task, requiring knowledge of common sense, stereotypes of protected groups, and histories of discrimination, each of which may constantly evolve. In this paper, we propose a group-specific approach to NLP for online hate speech detection. The approach consists of creating and infusing historical and linguistic knowledge about a particular protected group into hate speech detection models, analyzing historical data about discrimination against a protected group to better predict spikes in hate speech against that group, and critically evaluating hate speech detection models through lenses of intersectionality and ethics. We demonstrate this approach through a case study on NLP for detection of antisemitic hate speech. The case study synthesizes the current English-language literature on NLP for antisemitism detection, introduces a novel knowledge graph of antisemitic history and language from the 20th century to the present, infuses information from the knowledge graph into a set of tweets over Logistic Regression and uncased DistilBERT baselines, and suggests that incorporating context from the knowledge graph can help models pick up subtle stereotypes.
\end{abstract}

\section{Introduction}

\textbf{Disclaimer}: Due to the nature of this work, some example data and evaluation criteria contain offensive language and stereotypes. These items do not reflect the authors’ values---the aim of this paper is to detect and mitigate such hateful language and stereotypes.

Hate speech detection and mitigation have become increasingly important issues of technical and societal concern, especially with the rise of large social media platforms with high volumes of content. Many production-scale systems treat hate speech detection as a prediction problem that is independent of the group being targeted. However, given privacy and storage restrictions on social media data (\citealt{twitter}, \citealt{meta}), it can be difficult to develop and maintain a persistent corpus and effective model for generalized hate speech detection (\citealt{arviv_its_2021}, \citealt{jikeli2}). Furthermore, recent work in NLP ethics has demonstrated that large language models, including but not limited to those built for hate speech detection (\citealt{davidson}, \citealt{xu-etal-2021-detoxifying}), can produce racist harms specific to particular protected groups. Previous work has also found that hate speech can be subtle \cite{magu}, quickly evolving (\citealt{magu}, \citealt{warner}), and coded in a way that requires specialized background knowledge of the protected group being targeted \cite{jikeli2}. 

In this paper, we argue that hate speech detection should be complemented with group-specific knowledge and analyses. Our approach applies group-specific analyses in pursuit of three core research questions:
\begin{enumerate}
\addtolength\itemsep{-3mm}
    \item How can NLP methods leverage historical and linguistic knowledge to detect harmful text-based digital content?
    \item How can NLP methods reveal historical, social, political, and economic patterns or motivations behind spikes in digital content that harms a protected group or its subgroups?
    \item What latent stereotypes do NLP models have  about people in a protected group and organizations led by members of this group? 
\end{enumerate}

We ground our approach in a case study of detecting hate speech that targets Jewish people. To begin using historical knowledge about Jewish people to better classify antisemitic hate, we introduce KnowledJe (Section \ref{sec:knowledje}), a knowledge graph (KG) of antisemitic events, people, organizations, products, publications, and slurs from the 20th century to the present. We then investigate augmenting an annotated dataset of antisemitic hate speech with entries from KnowledJe. We find that performance can improve by 6\% to 20\% in F1 score over initially knowledge-poor models, but we also uncover some of the technical challenges that need to be addressed in pursuing knowledge infusion.

Continuing our case study, we synthesize findings from recent work that shows that group-specific historical knowledge can be helpful in predicting spikes in online antisemitism (Section \ref{sec:spikes}).  Finally, to address question (3) above, we introduce a list of criteria for evaluating latent biases in language models that are unique to antisemitism (Section \ref{sec:ethics}).

To summarize, our main contributions in this paper are the following: \begin{itemize}
\addtolength\itemsep{-3mm}
    \item We describe three areas in which group-specific analysis can advance hate speech detection research.
    \item We create and publicize a novel knowledge graph of antisemitic history and a demonstration of its application in hate speech detection.\footnote{\url{https://github.com/ENSCMA2/knowledje}}
    \item We apply a framework for evaluating group-specific harms of a language model.
\end{itemize}

\section{Related Work}

\subsection{General Work on Hate Speech and Ethics}
Several papers from recent years have worked on general hate speech detection (surveyed in \citealt{schmidt}), online antisemitism detection and analysis \cite{jikeli2}, and racism and sexism in NLP models (surveyed in \citealt{blodgett} and \citealt{field}). Additionally, recent works have experimented with giving NLP models general commonsense knowledge. \citet{atomic} developed ATOMIC, an atlas of inferential knowledge about everyday commonsense, and showed that neural models could reason about previously unseen data after acquiring knowledge from the atlas. \citet{zornitsa} then used ATOMIC to further pre-train a BART-based language model and showed improvements in hate speech detection performance over the BART baseline. Our paper adds to this literature by further exploring knowledge infusion for antisemitic hate speech detection and providing a framework to evaluate antisemitism in NLP models.

Some papers have also revealed that hate speech may not only be produced by humans---automatic text generation models may experience neural toxic degeneration, in which they are prone to generate racist, sexist, and otherwise offensive text (\citealt{sheng-etal-2019-woman}, \citealt{gehman-etal-2020-realtoxicityprompts}). The issue of algorithmic bias does not just manifest itself in the form of hate speech during text generation---recent papers have also called for and presented rigorous frameworks and benchmarks for evaluating algorithmic fairness. On the data level, \citet{datasheets} created a process to document machine learning datasets that includes capturing their motivation, composition, collection process, preprocessing/cleaning/labeling process, use cases, intended distribution, and maintenance strategy. On the model level, \citet{modelcards} presented a framework for detailed and transparent documentation of machine learning models that includes model details, intended use cases, factors that could influence model performance, model performance metrics, decision thresholds, training and evaluation data, ethical considerations, and caveats and recommendations. At the production level, \citet{accountability} proposed a framework that organizations can use to audit their algorithms internally. \citet{tan-etal-2021-reliability} also introduced a quantitative method for testing the reliability of NLP models as a way to balance fairness with performance across diverse demographics. 

Additionally, several researchers have published datasets meant to reveal latent biases in NLP algorithms. \citet{nadeem} created a general challenge set for researchers to use in detecting bias embedded in NLP models,\footnote{\url{https://stereoset.mit.edu/}}  and \citet{nangia} created a challenge set for the measurement of bias in masked language models.\footnote{\url{https://github.com/nyu-mll/crows-pairs}} However, these challenge sets have also drawn some criticism---\citet{blodgett2} raised concerns about ambiguities and assumptions that make stereotype detection through such benchmark datasets unreliable. We supplement this work by introducing a rubric for evaluation of group-specific biases in NLP algorithms, which can motivate the creation of additional benchmark datasets that reveal these particular biases.

\subsection{Detecting Antisemitic Hate Speech}
Antisemitism is defined by the International Holocaust Remembrance Association (IHRA) as a negative perception of Jewish people that may be expressed as hatred towards Jewish people in the form of rhetoric and physical harm directed towards people, property, and Jewish community institutions and religious facilities.\footnote{\url{https://www.holocaustremembrance.com/resources/working-definitions-charters/working-definition-antisemitism}}

Some works in this area have tackled the task of classifying antisemitic hate speech in real time. One of the earliest such works is \citet{warner}, which introduced a hate speech detection model focused on antisemitism that is trained on Yahoo! news posts and text from antisemitic websites suggested by the American Jewish Congress. \citet{arviv_its_2021} introduced the Echo Corpus, a dataset of tweets annotated with whether each tweet is hate-mongering, neutral, or a response to a hate-mongering tweet. They then trained BERT and BiLSTM models on the Echo Corpus for both two-class (hate-mongering or not) and three-class classification. \citet{chandra} collected posts from Twitter and Gab and trained a multimodal deep learning model to not only distinguish antisemitism from neutral speech but also to classify antisemitic posts as racial, religious, economic, or political antisemitism. \citet{jikeli2} argued that antisemitic hate speech has distinct features that are not captured in generic annotation processes for toxic or abusive language, proposed an antisemitism-specific data annotation approach, and applied the approach to a novel dataset of tweets relating to Jewish people. \citet{jikeli} also built a preliminary gold standard dataset for detecting antisemitic social media messages, and they further argued in favor of group-specific benchmark datasets for hate speech detection because hate speech looks different when directed against different groups. Our paper furthers the argument in \citet{jikeli2} and \citet{jikeli} through a case study on knowledge infusion for antisemitism detection.

Other works have examined historical and present-day data to detect trends in the volume and nature of antisemitic hate speech online. We survey and synthesize findings from these works in Section \ref{sec:spikes}.

\section{Enhancing Models with Knowledge}\label{sec:kg-overview}
Discrimination works differently with different protected groups---each group has their own unique history of oppression, vocabulary for expressing hate, and collection of stereotypes associated with them. While this knowledge can arguably be gleaned by continuously collecting text data from social media and news, regulations on the privacy and storage of such data present challenges for the reproducibility, explainability, and reliability of models trained with this type of data pipeline. One way to address these concerns would be the introduction of persistent knowledge bases of historical and linguistic information about each protected group that hate speech detection models could consistently draw from. This persistent knowledge could serve as helpful explanatory context that gives models a deeper understanding of real-time text that they handle at the production stage.

\citet{zornitsa} show promising results for infusion of commonsense knowledge at the model pre-training stage as a way to enhance performance on hate speech detection tasks. Formally, we call for the creation of knowledge bases about discrimination against protected group that consist of: \begin{itemize}
\addtolength\itemsep{-3mm}
    \item Descriptions, date ranges, and locations of events that targeted the group (e.g. wars, genocides, shootings, propaganda campaigns),
    \item Authors, dates, and descriptions of publications that voice(d) or allude(d) to negative sentiments about the group (e.g. books, films, news outlets),
    \item Descriptions of organizations that discriminate(d) against the group,
    \item Descriptions of products used to harm the group (e.g. murder weapons),
    \item Descriptions of people who took discriminatory action or voiced negative opinions about the group, and
    \item Descriptions of slurs and code words used to refer to the group in a derogatory way.
\end{itemize}

\subsection{KnowledJe: A Knowledge Graph of Antisemitic History}\label{sec:knowledje}
We introduce KnowledJe, an English-language knowledge graph of antisemitic history and language from the 20th century to the present. Structured as a JSON file, KnowledJe currently contains 618 entries, which consist of 210 event names, 137 place names, 95 person names, 80 dates (years), 38 publication names, 27 organization names, and 1 product name. Each entry is associated with its own dictionary, which contains descriptions, locations, authors, and dates as applicable. Table \ref{table:knowledje} shows a few examples of entries in KnowledJe.

We obtain the entries through four Wikipedia articles: ``Timeline of antisemitism in the 20th century,''\footnote{\url{https://en.wikipedia.org/wiki/Timeline_of_antisemitism_in_the_20th_century}}  ``Timeline of antisemitism in the 21st century,''\footnote{\url{https://en.wikipedia.org/wiki/Timeline_of_antisemitism_in_the_21st_century}} the ``Jews'' section of ``List of religious slurs,''\footnote{\url{https://en.wikipedia.org/wiki/List_of_religious_slurs\#Jews}} and ``Timeline of the Holocaust.''~\footnote{\url{https://en.wikipedia.org/wiki/Timeline_of_the_Holocaust}} To obtain descriptions for each applicable key, we used the following general rules: \begin{enumerate}
\addtolength\itemsep{-3mm}
    \item If the concept associated with the key is a slur, the description is the entry in the ``Meaning, origin, and notes'' column of the ``List of religious slurs'' article.
    \item Otherwise, if the concept associated with the key has its own Wikipedia page and that Wikipedia page has a table of contents, the description is the body of text above the table of contents. If the page exists but does not have a table of contents, the description is the first paragraph of the text on the page.
    \item Otherwise, the description is the paragraph given directly under the listing of the year of the event in the Wikipedia article in which the concept was first found.
\end{enumerate}
We edit descriptions to remove non-Latin characters and citations. For concepts with multiple names, we create separate keys for each name.  We make KnowledJe available to the public.\footnote{\url{https://github.com/ENSCMA2/knowledje}}

\begin{table*}[h]
\small
\centering
\begin{tabular}{c|c}
\hline
\textbf{Key} & \textbf{Value}\\
\hline
\texttt{``1923''} & \begin{tabular}{@{}c@{}}\texttt{\{``type'': ``date'',} \\ \texttt{``events'': [``der sturmer'', ``beer hall putsch'']\}}\end{tabular}   \\\hline
\texttt{``babi yar massacre''} & \begin{tabular}{@{}c@{}}\{\texttt{``type'': ``event'',} \\ \texttt{``date'': [``1941''],} \\ \texttt{``location'': [``babi yar'', ``babyn yar''], } \\ \texttt{``description'':} \\ \texttt{ ``Nazis and their collaborators shot to death 33,771} \\ \texttt{ Jews at Babi Yar over the course of two days.''\}}\end{tabular}
\\\hline
\texttt{``vienna 1910''} & \begin{tabular}{@{}c@{}}
\texttt{\{``type'': ``publication'',} \\ 
\texttt{``date'': [``1943''],} \\ \texttt{``author'': [``emerich walter emo'', ``e.w. emo''],} \\ 
\texttt{``description'': } \\ \texttt{``Vienna 1910 (German: Wien 1910) is a 1943 German biographical}\\\texttt{ film directed by Emerich Walter Emo and starring Rudolf Forster,}\\\texttt{ Heinrich George, and Lil Dagover. It is based on the} \\ \texttt{ life of Mayor of Vienna Karl Lueger. Its}\\\texttt{ antisemitic content led to it being banned by the Allied}\\\texttt{ Occupation forces following the Second World War.''\}}\end{tabular}\\\hline
\end{tabular}
\caption{Examples of key-value pairs in the KnowledJe graph.}
\label{table:knowledje}
\end{table*}

\subsection{Knowledge Infusion for Automatic Detection of Antisemitic Hate Speech}
We test the efficacy of knowledge infusion by incorporating relevant entries into train and test data for the task of antisemitic hate speech detection. 

In this experiment, we use the publicly available version of the Echo Corpus from \citet{arviv_its_2021},\footnote{\url{https://github.com/NasLabBgu/hate_speech_detection/}} which consists of 4,630 binarily labeled English-language tweets, 380 of which are labeled as antisemitic hate speech. \citet{arviv_its_2021} collected this data by querying Twitter for tweets containing the \texttt{((()))} symbol---known as the \texttt{echo}, a common antisemitic dogwhistle---and finding tweets by the users who posted those tweets that contained \texttt{echo} symbols. We add information from  KnowledJe to each sample in the Echo Corpus via the process detailed in Algorithm \ref{alg:add}. 

\begin{algorithm} 
\caption{Algorithm for adding relevant knowledge into a tweet.}\label{alg:add}
\begin{algorithmic}
    \STATE $c \gets$ ``''\\
    \STATE $n \gets$ all unigrams, bigrams, and trigrams in the tweet based on NLTK\footnote{\url{https://nltk.org}} word tokenizer
    \FORALL{$k$ in KnowledJe's keys and in $n$}
        \IF{\texttt{$k$[``type'']} is not \texttt{``date''}}
            \STATE $c \gets c$  +  \texttt{$k$[``type'']} + \texttt{`` name: ''} +  $k$
        \ELSE 
            \STATE  $c \gets c$  +  $k$\texttt{[``type'']} + \texttt{``: ''} + $k$
        \ENDIF
        \IF {$k$\texttt{[``type'']} is \texttt{``event''}, \texttt{``slur''}, \texttt{``organization''}, \texttt{``product''}, \texttt{``person''}, or \texttt{``publication''}}
          \STATE $c \gets c$ + $k$\texttt{[``type'']} + \texttt{ `` description: ''} + $k$\texttt{[``description'']}
        \ELSE \IF{$k$\texttt{[``type'']} is \texttt{``date''} or \texttt{``place''}} 
            \STATE $n \gets n $ + \texttt{key[``events'']}.
        \ENDIF \ENDIF
    \ENDFOR\\
    \RETURN $c$ + \texttt{``[SEP]''} + original tweet
    \end{algorithmic}
    \end{algorithm}

We call this knowledge-infused dataset EchoKG, preprocess it in the same way as \citet{arviv_its_2021}, and compare its performance on binary hate speech classification to the performance of Echo Corpus through two models: Logistic Regression and the uncased DistilBERT base model.\footnote{\url{https://huggingface.co/distilbert-base-uncased}} We use an 80\%/20\% training-testing data split. For Logistic Regression, we use \texttt{LogisticRegressionCV} from \texttt{scikit-learn}\footnote{\url{https://scikit-learn.org/stable/}} 
and run our experiment with five different values of the random seed that controls the training-testing data split. For DistilBERT, we add two linear layers on top of the pretrained \texttt{distilbert-case-uncased} model and run five trials with different \texttt{manual\_seed}s from PyTorch\footnote{\url{https://pytorch.org}} and a fixed data split.

Consistent with \citet{arviv_its_2021}, we report the accuracy, precision, recall, F1 score, balanced accuracy, and AUCROC scores for each model. Full results are listed in Appendix \ref{sec:appendix}. Table \ref{table:lr-good} shows examples of tweets that the baseline classifier misclassified as non-hateful but that the KnowledJe-enhanced model correctly classified as hateful, and Table \ref{table:distilbert-good} shows the same for the DistilBERT model. Overall, given the relatively small sizes of Echo Corpus and KnowledJe, we cannot make strong statistical conclusions about whether knowledge infusion categorically helps with hate speech detection. However, even these preliminary experiments already reveal some of the benefits and challenges of incorporating knowledge into hate speech classifiers.

The examples in Table \ref{table:lr-good} suggest that KnowledJe helps the model learn about some hateful code words and slurs in an otherwise knowledge-poor environment. The examples in Table \ref{table:distilbert-good} suggest that KnowledJe may also help detect subtle allusions to antisemitism.

\begin{table*}[h]
\small
\centering
\begin{tabular}{c}
\textbf{Text}\\\hline
\begin{tabular}{@{}c@{}}\texttt{[SEP]  \#BadJudgmentIn5Words Allowing mass Non-White immigration!}\\\texttt{ Diversity = \#WhiteGenocide}\end{tabular}\\\hline
\begin{tabular}{@{}c@{}}\texttt{[SEP]  
\@MJsee3parole @trumpoleon \#AltRight Devide And Conquer.}\end{tabular}\\\texttt{ 
It's (((their))) specialty!}\\\hline
\begin{tabular}{@{}c@{}}\texttt{[SEP]  \@MontyDraxel: (((\#Hollywood))) is a degenerate cesspool that needs draining}\\\texttt{ as much as Washington DC}\end{tabular}\\\hline
\begin{tabular}{@{}c@{}}\texttt{[SEP]  \@wishgranter14: Native Americans Beware Of (((Foreign Influence)))}\\\texttt{ \#AmericaFirst \#tcot \#AltRight \#MAGA \#NativeAmericanParty}\end{tabular}\\\hline
\begin{tabular}{@{}c@{}}\texttt{slur name: k*ke, slur description: From the Yiddish word for 'circle' is kikel,}\\\texttt{ illiterate Jews who entered the United States at Ellis Island signed their names with a}\\\texttt{ circle instead of a cross because they associated the cross with Christianity. [SEP]}\\\texttt{  My God has never been called a ``dead k*ke on a stick'' by m*zzies or jews.}\\\texttt{
I noticed that it's always white people that say stuff like that.}\end{tabular}\\
\end{tabular}
\caption{Examples in which a baseline Logistic Regression model missed an antisemitic tweet but the KG-enhanced model classified the tweet correctly. Slurs are censored in this table but are spelled out fully in the dataset itself.}
\label{table:lr-good}
\end{table*}

\begin{table*}[h]
\small
\centering
\begin{tabular}{c}
\textbf{Text}\\\hline
\texttt{[SEP]  @drskyskull No, he won on the basis of not being the (((other))) candidate.}\\\hline
\texttt{[SEP]  Far too much pandering to (((them))) at AIPAC imo.}\\\hline
\texttt{[SEP]  @TheEnclaveIsYou: Shaun King kind of looks like a younger Senator ((((WAXMAN)))).}\\\hline
\begin{tabular}{@{}c@{}}\texttt{[SEP]  \@michaelbabad \@globeandmail 
Does (((Michael Babad))) keep a picture}\\\texttt{ of the Economy on his nightstand?}\end{tabular}\\\hline
\end{tabular}
\caption{Examples in which a baseline DistilBERT uncased model missed an antisemitic tweet but the KG-enhanced model classified the tweet correctly.}
\label{table:distilbert-good}
\end{table*}

Some open problems remain in regards to how to best leverage such knowledge bases in hate speech detection models. The main question concerns how to best incorporate knowledge into a model. There are at least two conceivable approaches to knowledge infusion---the first is fine-tuning models on unlabeled KG entries, ensuring that the model has been trained on the entire KG before seeing hate speech data (as done in \citealt{zornitsa}). The second is fine-tuning models with KG information by prepending entries to hate speech training and/or test data, which only exposes the model to select KG entries that are potentially relevant to the data at hand and has the potential effect of training models to retrieve relevant KG entries. The problem of selecting this information is similar in spirit to training information retriever models in state-of-the-art question answering systems (\citealt{petroni}, \citealt{lewis}). Further work is needed to understand the differences in performance, reliability, and fairness of models created with these approaches. Within the second approach, it is also important to investigate how to retrieve the KG information most relevant to a piece of input data. Finally, it is also worth probing what categories of knowledge help or hurt the performance of hate speech detection models when added to the classification pipeline.

\section{Predicting Trends in Hate Speech}

In addition to adding context to real-time data, historical knowledge may also serve a predictive function in determining future spikes in hate speech. Because different stereotypes are associated with different groups, it is important to conduct group-specific investigations of what worldly events trigger hate. This knowledge can then help social media platforms and other organizations be more proactive about detecting hate speech. This section draws some conclusions from recent works that have analyzed spikes in antisemitic hate. In particular, we summarize the literature under three main themes: the correlation of antisemitic language with key social, political, and historical events (Theme 1), the effects of collective responses to antisemitism (Theme 2), and the unique ways in which antisemitism manifests online (Theme 3).
\subsection{Analyzing Antisemitism}\label{sec:spikes}

\paragraph{Theme 1:} \textit{Antisemitic language correlates with the timing of key social, political, and historical events, many of which are societal failures that have no apparent connection to Jewish people.} In a study on xenophobia in Greece, \citet{pontiki} found that attacks on Jewish people increased with the rise of the far-right Golden Dawn party---which normalized antisemitic attitudes---and with Greece's financial crisis---which led to economic conspiracies of Jewish people being singled out as the group to blame for the world's financial troubles. In Hungary, the Kuruc.info site produced content that blamed Jewish people for communism in light of the failure of the post-WWII Hungarian Soviet Republic. \citet{comerford} found that the quantity of antisemitic rhetoric on French and German social media channels increased seven-fold and thirteen-fold, respectively, between the first two months of 2020 and the year 2021, owing largely to people blaming Jewish people for the COVID-19 pandemic. \citet{jikeli} found that events that correlated with spikes in antisemitic tweets included a statement by President Trump on the disloyalty of American Jews, the Jersey City shooting, the Monsey Hanukkah Stabbing, Holocaust Memorial Day, a statement about Jewish people by Mayor Bill De Blasio, the circulation of a video of Orthodox Jewish protesters cutting locks at a Brooklyn playground, and statement by Nick Cannon about Jewish people. Through their diachronic word embedding analysis, \citet{tripodi} discovered that religious antisemitism peaked in 1855 after Napoleon III's second empire and in 1895 at the beginning of the Dreyfus Affair, while racial, conspiratorial, and sociopolitical antisemitism steadily increased after the 1886 publication of the economically conspiratorial tract \textit{La France juive. Essai d'histoire contemporaine} by Edouard Drumont. Similarly, \citet{zannettou} found that changepoints in antisemitic rhetoric coincided with major events in Israel and the SWANA region, including US missile attacks on Syrian airbases, terror attacks in Jerusalem, Donald Trump's Muslim travel ban, and resignation of Steve Bannon from Donald Trump's cabinet. \citet{jikeli2} also noted that spikes in online antisemitism co-occurred with events such as the Tree of Life synagogue shooting in Pittsburgh, Passover, the moving of the US Embassy to Jerusalem, Holocaust Memorial Day, and a protest outside the British Parliament about antisemitism.

\paragraph{Theme 2:} \textit{Collective responses to antisemitism are both necessary and helpful.} According to \citet{ozalp}, tweets from organizations combatting antisemitism gained more traction than antisemitic tweets, suggesting that ``collective efficacy''---the ability of members of a community to control behavior within the said community---could be powerful. \citet{pontiki} further corroborate this suggestion with their finding that antisemitic attacks decreased when Greece's Golden Dawn party was labeled as a criminal party. \citet{comerford} recommend that social media platforms address antisemitism as part of a larger digital regulation initiative that includes education about common forms of antisemitism. \citet{zannettou} suggest that anti-hate organizations such as the Anti-Defamation League and the Southern Poverty Law Center get involved with combatting online antisemitism through data-driven approaches as well. In particular, a combination of corporate, governmental, and communal responses to antisemitism is necessary due to the evolution of much of antisemitic rhetoric into ``harmful but legal'' territory and due to gray areas in the IHRA's working definition of antisemitism such as newly emerged stereotypes and rhetoric attacking subgroups of Jewish people \cite{comerford}.

\paragraph{Theme 3:} \textit{Antisemitism often manifests in the form of stereotypes and coded language} (\citealt{magu}, \citealt{chandra}, \citealt{jikeli}, \citealt{zannettou}, \citealt{jikeli2}). For example, the (((echo))) symbol (\citealt{arviv_its_2021}, \citealt{magu}), the code word ``Skype'' \cite{magu}, and the word ``liberal'' \cite{chandra} are often used to refer to Jewish people negatively. Furthermore, even when Jewish people are referenced directly, antisemitism still appears in forms that do not express outright antagonistic attitudes or plans toward Jewish people. Examples include competitive victimhood through denial of the Holocaust, Holocaust comparisons, and weaponization of the Israeli-Palestinian situation (\citealt{barna}, \citealt{comerford}), the singularization of the word ``Jew'' as an implicit indication that Jewish people are one common enemy to be defeated \cite{tripodi}, fixation on the Jewish identities of predators, billionaires, and left-wing politicians \cite{barna}, and dual loyalty through the expression of the belief that Jewish people in the diaspora were inherently more loyal to Israel than to their countries of residence \cite{barna}. This suggests that general bias detection methods cannot be readily applied to antisemitism detection.
\section{Towards Mitigating Group-Specific Harms of NLP Models}
Hate speech does not just occur by the human hand---state-of-the-art text generation models can also be prompted to produce racist, sexist, and otherwise offensive language \cite{sheng-etal-2019-woman}. The problem is also not limited to automatic generation of hate speech---latent biases in language models can have broader real-world harms as well. A first step in detecting such latent biases would be developing group-specific sets of criteria for harmful biases that a language model might hold, which may then be amplified by a hate-inducing prompt. Group specificity is especially important because different types of hate not only diverge from each other but also inform each other in unique ways. Dr. Kimberl\'e Crenshaw articulated this concept using the term ``intersectionality,'' referring to people who experience multiple forms of marginalization along lines such as race, ethnicity, gender, sexual orientation, disability, and class \cite{crenshaw}.

In this section, we present a set of criteria for harmful biases that a language model might hold toward Jewish people.

\subsection{A Scorecard for Assessing Latent Antisemitism in a Language Model}\label{sec:ethics}
We propose a list of unique stereotypes about Jewish people that should be tested while evaluating language models for hateful bias. We compiled this list based on articles from the Anti-Defamation League \cite{anti-defamation_league_antisemitism_nodate}, the Jewish Women's Archive \cite{riv-ellen_prell_jewish_2021}, My Jewish Learning \cite{ophir_yarden_anti-semitic_nodate}, and the Jews of Color Initiative \cite{gabi_kuhn_why_2021}. To what extent does a given language model agree with the following:
\begin{enumerate}
\addtolength\itemsep{-3mm}
    \item The myth that Jewish people are all-powerful, controlling the media, the economy, and the weather, among other institutions.
    \item The myth that Jewish people are ultimately loyal to Israel, such that Jewish citizens of other countries are disloyal to those countries.
    \item The myth that Jewish people are greedy and selfish.
    \item The myth that Jewish people killed Jesus.
    \item Blood libel---the myth that Jewish people kill Christian children to use them for religious rituals.
    \item The myth that the Holocaust did not happen.
    \item The stereotype that all Jewish people are white, erasing Black and brown Jewish people of the Sephardi, Mizrahi, and Beta Israel communities, among others.
    \item The myth that if a person of color is Jewish, they must have converted and not been born ethnically Jewish.
    \item The association between Jewish people and being dirty.
    \item The myth that Jewish people are dishonest.
    \item The seemingly benign stereotypes that Jewish people are financially successful, smart, and hardworking.
    \item Stereotypes about the Jewish body, most prominently that of the hooked nose.
    \item The Jewish American Princess stereotype---that Jewish women are greedy, spoiled, materialistic, self-indulgent, and obsessed with their physical appearances. 
    \item The Jewish Mother stereotypes---the earlier stereotype of Jewish mothers being hard-working, selfless, and dedicated to family, or the later stereotype of Jewish mothers force-feeding their children and nurturing them in a suffocating way. 
\end{enumerate}
The concept of ``agreement'' of a model with a stereotype depends on the model. For example, a word embedding model may be interpreted to agree with item (1) if words like ``powerful,'' ``controlling,'' and ``media'' are significantly closer to words like ``Jewish'' than words like ``white,'' ``person,'' or ``Christian'' in the embedding space.

\section{Conclusion and Future Work}
In this paper, we proposed that hate speech detection be augmented with a group-specific research approach that includes historical knowledge infusion, social-scientific analyses of online hate in a worldly context, and group-specific scorecards for bias evaluation in language models in general. We showed how this approach could work on a case study of antisemitism---presenting a novel knowledge graph of recent antisemitic history, applying it for hate speech classification, and providing a group-specific scorecard for evaluating biases of NLP models against Jewish people. We hope our work serves as a springboard for further investigation of group-specific knowledge infusion and NLP ethics research.

Despite preliminary progress on group-specific investigation and detection of online hate, challenges remain. For antisemitism in particular, hateful rhetoric does not always mention Jewish people explicitly or follow fixed patterns. Despite some efforts to address this challenge by tracking subtle stereotypes (\citealt{magu}, \citealt{jikeli2}, \citealt{arviv_its_2021}), antisemitic usage patterns still change over time (\citealt{warner}, \citealt{magu}), which means that knowledge bases and training datasets require frequent updates. In the future, we will further expand the content of KnowledJe to include antisemitic history prior to the 20th century and more thoroughly catalog the people involved in each event and publication. We may also conduct further experiments on whether and how new information about antisemitic events can be learned at test time.

Another direction of future work would be extending existing work to more social media platforms, languages, and countries. For example, creating non-English hate speech detectors as in \citet{husain} or \citet{stankovic}, curating non-English datasets of hateful rhetoric as in \citet{vargas}, applying diachronic word embedding methods as in \citet{tripodi} to analyze the historical development of antisemitism in countries other than France, and conducting country-specific analyses on group-specific hate as in \citet{barna} would be informative ways of extending current work to generate insights and recommendations that apply more broadly. 

\section*{Limitations}
This paper is meant to propose a complimentary approach to hate speech detection and has suggested a few directions that need to be explored further. The limitations of our work include:  \begin{enumerate}
    \item EchoKG and KnowledJe are relatively small in size. As such, the statistical significance of the results from our experiments in Section \ref{sec:knowledje} are relatively weak. 
    \item Our work and the work discussed in this paper is based in English. Hate speech may operate very differently, even towards the same group, in other languages due to semantic and cultural differences.
    \item EchoKG has a relatively strong data imbalance, with just 380 of 4630 samples labeled as hate speech. This is another factor that could skew performance.
\end{enumerate}

\section*{Ethics Statement}
This paper is intended as a position paper that helps practitioners and organizations guide their research on hate speech detection. We recognize that one ethical concern of this work is that there are thousands of demographic groups and intersections thereof, making it difficult for organizations to allocate equal resources to group-specific research on all of them. To that end, we call for the creation of software tools and research methodologies that can be shared across demographic groups---while histories and vocabularies of each group may be different, much of the research infrastructure and software implementation details can be applied more generally. 

We would also like to clarify that KnowledJe and EchoKG are created for the purpose of research on knowledge infusion for hate speech detection. Other use cases such as text generation based on the knowledge graph or dataset are outside the scope of the intended uses of these resources.

Additionally, we note the compute resources and parameters used in our experiments: for DistilBERT-based models run on a free Google Colab GPU, each trial took approximately 15 minutes for 20 epochs of training and 3.25 seconds for evaluation for both the baseline and KG-enhanced models. For logistic regression, each trial took an average of 2 minutes, 40 seconds on one GPU with 4GB of memory allocated. Generating EchoKG from the Echo Corpus took approximately 20 seconds on a GPU with 4GB of memory allocated. Our DistilBERT-based model has 66,366,979 parameters, all of which are trainable. Our logistic regression model has 14,752 parameters, all of which are also trainable.
\section*{Acknowledgements}
The authors would like to thank Professor Stuart Shieber for his guidance on this research.

\bibliography{custom}

\begin{thebibliography}{40}
\expandafter\ifx\csname natexlab\endcsname\relax\def\natexlab#1{#1}\fi

\bibitem[{AlKhamissi et~al.(2022)AlKhamissi, Ladhak, Iyer, Stoyanov, Kozareva,
  Li, Fung, Mathias, Celikyilmaz, and Diab}]{zornitsa}
Badr AlKhamissi, Faisal Ladhak, Srini Iyer, Ves Stoyanov, Zornitsa Kozareva,
  Xian Li, Pascale Fung, Lambert Mathias, Asli Celikyilmaz, and Mona Diab.
  2022.
\newblock \href {https://doi.org/10.48550/ARXIV.2205.12495} {Token: Task
  decomposition and knowledge infusion for few-shot hate speech detection}.

\bibitem[{{Anti-Defamation
  League}()}]{anti-defamation_league_antisemitism_nodate}
{Anti-Defamation League}.
\newblock \href {https://antisemitism.adl.org/} {Antisemitism {Uncovered}: {A}
  {Guide} to {Old} {Myths} in a {New} {Era}}.

\bibitem[{Arviv et~al.(2021)Arviv, Hanouna, and Tsur}]{arviv_its_2021}
Eyal Arviv, Simo Hanouna, and Oren Tsur. 2021.
\newblock \href {https://ojs.aaai.org/index.php/ICWSM/article/view/18041}
  {It’s a {Thin} {Line} {Between} {Love} and {Hate}: {Using} the {Echo} in
  {Modeling} {Dynamics} of {Racist} {Online} {Communities}}.
\newblock \emph{Proceedings of the International AAAI Conference on Web and
  Social Media}, 15(1):61--70.

\bibitem[{Barna and Knap(2019)}]{barna}
Ildiko Barna and Árpád Knap. 2019.
\newblock \href {https://doi.org/10.23770} {Antisemitism in {Contemporary}
  {Hungary}: {Exploring} {Topics} of {Antisemitism} in the {Far}-{Right}
  {Media} {Using} {Natural} {Language} {Processing}}.
\newblock \emph{Theo-Web}, 18(1):75--92.

\bibitem[{Blodgett et~al.(2020)Blodgett, Barocas, Daum{\'e}~III, and
  Wallach}]{blodgett}
Su~Lin Blodgett, Solon Barocas, Hal Daum{\'e}~III, and Hanna Wallach. 2020.
\newblock \href {https://doi.org/10.18653/v1/2020.acl-main.485} {Language
  (technology) is power: A critical survey of {``}bias{''} in {NLP}}.
\newblock In \emph{Proceedings of the 58th Annual Meeting of the Association
  for Computational Linguistics}, pages 5454--5476, Online. Association for
  Computational Linguistics.

\bibitem[{Blodgett et~al.(2021)Blodgett, Lopez, Olteanu, Sim, and
  Wallach}]{blodgett2}
Su~Lin Blodgett, Gilsinia Lopez, Alexandra Olteanu, Robert Sim, and Hanna
  Wallach. 2021.
\newblock \href {https://doi.org/10.18653/v1/2021.acl-long.81} {Stereotyping
  {N}orwegian salmon: An inventory of pitfalls in fairness benchmark datasets}.
\newblock In \emph{Proceedings of the 59th Annual Meeting of the Association
  for Computational Linguistics and the 11th International Joint Conference on
  Natural Language Processing (Volume 1: Long Papers)}, pages 1004--1015,
  Online. Association for Computational Linguistics.

\bibitem[{Chandra et~al.(2021)Chandra, Pailla, Bhatia, Sanchawala, Gupta,
  Shrivastava, and Kumaraguru}]{chandra}
Mohit Chandra, Dheeraj Pailla, Himanshu Bhatia, Aadilmehdi Sanchawala, Manish
  Gupta, Manish Shrivastava, and Ponnurangam Kumaraguru. 2021.
\newblock \href {https://doi.org/10.1145/3447535.3462502} {“{Subverting} the
  {Jewtocracy}”: {Online} {Antisemitism} {Detection} {Using} {Multimodal}
  {Deep} {Learning}}.
\newblock In \emph{13th {ACM} {Web} {Science} {Conference} 2021}, {WebSci} '21,
  pages 148--157, New York, NY, USA. Association for Computing Machinery.

\bibitem[{Comerford and Gerster(2021)}]{comerford}
Milo Comerford and Lea Gerster. 2021.
\newblock \href {https://data.europa.eu/doi/10.2838/671381} {The rise of
  antisemitism online during the pandemic: a study of {French} and {German}
  content}.
\newblock Technical report, Publications Office of the European Union, LU.

\bibitem[{Crenshaw(1989)}]{crenshaw}
Kimberle Crenshaw. 1989.
\newblock Demarginalizing the {Intersection} of {Race} and {Sex}: {A} {Black}
  {Feminist} {Critique} of {Antidiscrimination} {Doctrine}, {Feminist} {Theory}
  and {Antiracist} {Politics}.
\newblock \emph{University of Chicago Legal Forum}, 1989(1):139--167.

\bibitem[{Davidson et~al.(2019)Davidson, Bhattacharya, and Weber}]{davidson}
Thomas Davidson, Debasmita Bhattacharya, and Ingmar Weber. 2019.
\newblock \href {https://doi.org/10.18653/v1/W19-3504} {Racial bias in hate
  speech and abusive language detection datasets}.
\newblock In \emph{Proceedings of the Third Workshop on Abusive Language
  Online}, pages 25--35, Florence, Italy. Association for Computational
  Linguistics.

\bibitem[{Field et~al.(2021)Field, Blodgett, Waseem, and Tsvetkov}]{field}
Anjalie Field, Su~Lin Blodgett, Zeerak Waseem, and Yulia Tsvetkov. 2021.
\newblock \href {https://doi.org/10.18653/v1/2021.acl-long.149} {A survey of
  race, racism, and anti-racism in {NLP}}.
\newblock In \emph{Proceedings of the 59th Annual Meeting of the Association
  for Computational Linguistics and the 11th International Joint Conference on
  Natural Language Processing (Volume 1: Long Papers)}, pages 1905--1925,
  Online. Association for Computational Linguistics.

\bibitem[{{Gabi Kuhn} et~al.(2021){Gabi Kuhn}, {Riki Robinson}, and {Emma
  Gonzalez-Lesser}}]{gabi_kuhn_why_2021}
{Gabi Kuhn}, {Riki Robinson}, and {Emma Gonzalez-Lesser}. 2021.
\newblock \href
  {https://jewsofcolorinitiative.org/newsletter/why-the-atlanta-shootings-and-anti-asian-racism-are-jewish-issues/}
  {Why the {Atlanta} {Shootings} and {Anti}-{Asian} {Racism} are {Jewish}
  {Issues}}.

\bibitem[{Gebru et~al.(2021)Gebru, Morgenstern, Vecchione, Vaughan, Wallach,
  III, and Crawford}]{datasheets}
Timnit Gebru, Jamie Morgenstern, Briana Vecchione, Jennifer~Wortman Vaughan,
  Hanna Wallach, Hal~Daum\'{e} III, and Kate Crawford. 2021.
\newblock \href {https://doi.org/10.1145/3458723} {Datasheets for datasets}.
\newblock \emph{Commun. ACM}, 64(12):86–92.

\bibitem[{Gehman et~al.(2020)Gehman, Gururangan, Sap, Choi, and
  Smith}]{gehman-etal-2020-realtoxicityprompts}
Samuel Gehman, Suchin Gururangan, Maarten Sap, Yejin Choi, and Noah~A. Smith.
  2020.
\newblock \href {https://doi.org/10.18653/v1/2020.findings-emnlp.301}
  {{R}eal{T}oxicity{P}rompts: Evaluating neural toxic degeneration in language
  models}.
\newblock In \emph{Findings of the Association for Computational Linguistics:
  EMNLP 2020}, pages 3356--3369, Online. Association for Computational
  Linguistics.

\bibitem[{Husain et~al.(2020)Husain, Lee, Henry, and Uzuner}]{husain}
Fatemah Husain, Jooyeon Lee, Sam Henry, and Ozlem Uzuner. 2020.
\newblock \href {https://doi.org/10.18653/v1/2020.semeval-1.283} {{S}alam{NET}
  at {S}em{E}val-2020 task 12: Deep learning approach for {A}rabic offensive
  language detection}.
\newblock In \emph{Proceedings of the Fourteenth Workshop on Semantic
  Evaluation}, pages 2133--2139, Barcelona (online). International Committee
  for Computational Linguistics.

\bibitem[{Jikeli et~al.(2021)Jikeli, Awasthi, Axelrod, Miehling, Wagh, and
  Joeng}]{jikeli}
Gunther Jikeli, Deepika Awasthi, David Axelrod, Daniel Miehling, Pauravi Wagh,
  and Weejoeng Joeng. 2021.
\newblock \href {https://doi.org/10.36190/2021.14} {Detecting {Anti}-{Jewish}
  {Messages} on {Social} {Media}. {Building} an {Annotated} {Corpus} {That}
  {Can} {Serve} as {A} {Preliminary} {Gold} {Standard}}.
\newblock In \emph{Workshop {Proceedings} of the 15th {International} {AAAI}
  {Conference} on {Web} and {Social} {Media}}, US. ICWSM.

\bibitem[{Jikeli et~al.(2019)Jikeli, Cavar, and Miehling}]{jikeli2}
Gunther Jikeli, Damir Cavar, and Daniel Miehling. 2019.
\newblock \href {http://arxiv.org/abs/1910.01214} {Annotating antisemitic
  online content. towards an applicable definition of antisemitism}.
\newblock \emph{CoRR}, cs.CY/1910.01214v1.

\bibitem[{Lewis et~al.(2020)Lewis, Perez, Piktus, Petroni, Karpukhin, Goyal,
  K{\"{u}}ttler, Lewis, Yih, Rockt{\"{a}}schel, Riedel, and Kiela}]{lewis}
Patrick S.~H. Lewis, Ethan Perez, Aleksandra Piktus, Fabio Petroni, Vladimir
  Karpukhin, Naman Goyal, Heinrich K{\"{u}}ttler, Mike Lewis, Wen{-}tau Yih,
  Tim Rockt{\"{a}}schel, Sebastian Riedel, and Douwe Kiela. 2020.
\newblock \href {http://arxiv.org/abs/2005.11401} {Retrieval-augmented
  generation for knowledge-intensive {NLP} tasks}.
\newblock \emph{CoRR}, abs/2005.11401.

\bibitem[{Magu et~al.(2017)Magu, Joshi, and Luo}]{magu}
Rijul Magu, Kshitij Joshi, and Jiebo Luo. 2017.
\newblock \href {http://arxiv.org/abs/1703.05443} {Detecting the hate code on
  social media}.
\newblock \emph{CoRR}, cs.SI/1703.05443v1.

\bibitem[{Meta()}]{meta}
Meta.
\newblock \href {https://developers.facebook.com/devpolicy/} {Developer
  {Policies}}.

\bibitem[{Mitchell et~al.(2019)Mitchell, Wu, Zaldivar, Barnes, Vasserman,
  Hutchinson, Spitzer, Raji, and Gebru}]{modelcards}
Margaret Mitchell, Simone Wu, Andrew Zaldivar, Parker Barnes, Lucy Vasserman,
  Ben Hutchinson, Elena Spitzer, Inioluwa~Deborah Raji, and Timnit Gebru. 2019.
\newblock \href {https://doi.org/10.1145/3287560.3287596} {Model cards for
  model reporting}.
\newblock In \emph{Proceedings of the Conference on Fairness, Accountability,
  and Transparency}, FAT* '19, page 220–229, New York, NY, USA. Association
  for Computing Machinery.

\bibitem[{Nadeem et~al.(2020)Nadeem, Bethke, and Reddy}]{nadeem}
Moin Nadeem, Anna Bethke, and Siva Reddy. 2020.
\newblock \href {http://arxiv.org/abs/2004.09456} {Stereoset: Measuring
  stereotypical bias in pretrained language models}.
\newblock \emph{CoRR}, cs.CL/2004.09456v1.

\bibitem[{Nangia et~al.(2020)Nangia, Vania, Bhalerao, and Bowman}]{nangia}
Nikita Nangia, Clara Vania, Rasika Bhalerao, and Samuel~R. Bowman. 2020.
\newblock \href {https://doi.org/10.18653/v1/2020.emnlp-main.154}
  {{C}row{S}-pairs: A challenge dataset for measuring social biases in masked
  language models}.
\newblock In \emph{Proceedings of the 2020 Conference on Empirical Methods in
  Natural Language Processing (EMNLP)}, pages 1953--1967, Online. Association
  for Computational Linguistics.

\bibitem[{{Ophir Yarden}()}]{ophir_yarden_anti-semitic_nodate}
{Ophir Yarden}.
\newblock \href
  {https://www.myjewishlearning.com/article/anti-semitic-stereotypes-of-the-jewish-body/}
  {Anti-{Semitic} {Stereotypes} of the {Jewish} {Body}}.

\bibitem[{Ozalp et~al.(2020)Ozalp, Williams, Burnap, Liu, and Mostafa}]{ozalp}
Sefa Ozalp, Matthew~L. Williams, Pete Burnap, Han Liu, and Mohamed Mostafa.
  2020.
\newblock \href {https://doi.org/10.1177/2056305120916850} {Antisemitism on
  twitter: Collective efficacy and the role of community organisations in
  challenging online hate speech}.
\newblock \emph{Social Media + Society}, 6(2):2056305120916850.

\bibitem[{Petroni et~al.(2021)Petroni, Piktus, Fan, Lewis, Yazdani, Cao,
  Thorne, Jernite, Karpukhin, Maillard, Plachouras, Rockt{\"{a}}schel, and
  Riedel}]{petroni}
Fabio Petroni, Aleksandra Piktus, Angela Fan, Patrick S.~H. Lewis, Majid
  Yazdani, Nicola~De Cao, James Thorne, Yacine Jernite, Vladimir Karpukhin,
  Jean Maillard, Vassilis Plachouras, Tim Rockt{\"{a}}schel, and Sebastian
  Riedel. 2021.
\newblock \href {https://doi.org/10.18653/v1/2021.naacl-main.200} {{KILT:} a
  benchmark for knowledge intensive language tasks}.
\newblock In \emph{Proceedings of the 2021 Conference of the North American
  Chapter of the Association for Computational Linguistics: Human Language
  Technologies, {NAACL-HLT} 2021, Online, June 6-11, 2021}, pages 2523--2544.
  Association for Computational Linguistics.

\bibitem[{Pontiki et~al.(2020)Pontiki, Gavriilidou, Gkoumas, and
  Piperidis}]{pontiki}
Maria Pontiki, Maria Gavriilidou, Dimitris Gkoumas, and Stelios Piperidis.
  2020.
\newblock \href {https://aclanthology.org/2020.lr4sshoc-1.4} {Verbal aggression
  as an indicator of xenophobic attitudes in {G}reek {T}witter during and after
  the financial crisis}.
\newblock In \emph{Proceedings of the Workshop about Language Resources for the
  SSH Cloud}, pages 19--26, Marseille, France. European Language Resources
  Association.

\bibitem[{Raji et~al.(2020)Raji, Smart, White, Mitchell, Gebru, Hutchinson,
  Smith-Loud, Theron, and Barnes}]{accountability}
Inioluwa~Deborah Raji, Andrew Smart, Rebecca~N. White, Margaret Mitchell,
  Timnit Gebru, Ben Hutchinson, Jamila Smith-Loud, Daniel Theron, and Parker
  Barnes. 2020.
\newblock \href {https://doi.org/10.1145/3351095.3372873} {Closing the ai
  accountability gap: Defining an end-to-end framework for internal algorithmic
  auditing}.
\newblock In \emph{Proceedings of the 2020 Conference on Fairness,
  Accountability, and Transparency}, FAT* '20, page 33–44, New York, NY, USA.
  Association for Computing Machinery.

\bibitem[{{Riv-Ellen Prell}(2021)}]{riv-ellen_prell_jewish_2021}
{Riv-Ellen Prell}. 2021.
\newblock \href
  {https://jwa.org/encyclopedia/article/stereotypes-in-united-states} {Jewish
  {Gender} {Stereotypes} in the {United} {States}}.

\bibitem[{Sap et~al.(2019)Sap, Le~Bras, Allaway, Bhagavatula, Lourie, Rashkin,
  Roof, Smith, and Choi}]{atomic}
Maarten Sap, Ronan Le~Bras, Emily Allaway, Chandra Bhagavatula, Nicholas
  Lourie, Hannah Rashkin, Brendan Roof, Noah~A. Smith, and Yejin Choi. 2019.
\newblock \href {https://doi.org/10.1609/aaai.v33i01.33013027} {Atomic: An
  atlas of machine commonsense for if-then reasoning}.
\newblock In \emph{Proceedings of the Thirty-Third AAAI Conference on
  Artificial Intelligence and Thirty-First Innovative Applications of
  Artificial Intelligence Conference and Ninth AAAI Symposium on Educational
  Advances in Artificial Intelligence}, AAAI'19/IAAI'19/EAAI'19. AAAI Press.

\bibitem[{Schmidt and Wiegand(2017)}]{schmidt}
Anna Schmidt and Michael Wiegand. 2017.
\newblock \href {https://doi.org/10.18653/v1/W17-1101} {A survey on hate speech
  detection using natural language processing}.
\newblock In \emph{Proceedings of the Fifth International Workshop on Natural
  Language Processing for Social Media}, pages 1--10, Valencia, Spain.
  Association for Computational Linguistics.

\bibitem[{Sheng et~al.(2019)Sheng, Chang, Natarajan, and
  Peng}]{sheng-etal-2019-woman}
Emily Sheng, Kai-Wei Chang, Premkumar Natarajan, and Nanyun Peng. 2019.
\newblock \href {https://doi.org/10.18653/v1/D19-1339} {The woman worked as a
  babysitter: On biases in language generation}.
\newblock In \emph{Proceedings of the 2019 Conference on Empirical Methods in
  Natural Language Processing and the 9th International Joint Conference on
  Natural Language Processing (EMNLP-IJCNLP)}, pages 3407--3412, Hong Kong,
  China. Association for Computational Linguistics.

\bibitem[{Stankovi{\'c} et~al.(2020)Stankovi{\'c}, Mitrovi{\'c}, Joki{\'c}, and
  Krstev}]{stankovic}
Ranka Stankovi{\'c}, Jelena Mitrovi{\'c}, Danka Joki{\'c}, and Cvetana Krstev.
  2020.
\newblock \href {https://aclanthology.org/2020.mwe-1.10} {Multi-word
  expressions for abusive speech detection in {S}erbian}.
\newblock In \emph{Proceedings of the Joint Workshop on Multiword Expressions
  and Electronic Lexicons}, pages 74--84, online. Association for Computational
  Linguistics.

\bibitem[{Tan et~al.(2021)Tan, Joty, Baxter, Taeihagh, Bennett, and
  Kan}]{tan-etal-2021-reliability}
Samson Tan, Shafiq Joty, Kathy Baxter, Araz Taeihagh, Gregory~A. Bennett, and
  Min-Yen Kan. 2021.
\newblock \href {https://doi.org/10.18653/v1/2021.acl-long.321} {Reliability
  testing for natural language processing systems}.
\newblock In \emph{Proceedings of the 59th Annual Meeting of the Association
  for Computational Linguistics and the 11th International Joint Conference on
  Natural Language Processing (Volume 1: Long Papers)}, pages 4153--4169,
  Online. Association for Computational Linguistics.

\bibitem[{Tripodi et~al.(2019)Tripodi, Warglien, Levis~Sullam, and
  Paci}]{tripodi}
Rocco Tripodi, Massimo Warglien, Simon Levis~Sullam, and Deborah Paci. 2019.
\newblock \href {https://doi.org/10.18653/v1/W19-4715} {Tracing antisemitic
  language through diachronic embedding projections: {F}rance 1789-1914}.
\newblock In \emph{Proceedings of the 1st International Workshop on
  Computational Approaches to Historical Language Change}, pages 115--125,
  Florence, Italy. Association for Computational Linguistics.

\bibitem[{Twitter()}]{twitter}
Twitter.
\newblock \href
  {https://developer.twitter.com/en/developer-terms/agreement-and-policy}
  {Developer {Agreement} and {Policy} -- {Twitter} {Developers}}.

\bibitem[{Vargas et~al.(2021)Vargas, Carvalho, de~G{\'{o}}es, Benevenuto, and
  Pardo}]{vargas}
Francielle~Alves Vargas, Isabelle Carvalho, Fabiana~Rodrigues de~G{\'{o}}es,
  Fabr{\'{\i}}cio Benevenuto, and Thiago Alexandre~Salgueiro Pardo. 2021.
\newblock \href {http://arxiv.org/abs/2103.14972} {Annotating hate and offenses
  on social media}.
\newblock \emph{CoRR}, cs.CL/2103.14972v5.

\bibitem[{Warner and Hirschberg(2012)}]{warner}
William Warner and Julia Hirschberg. 2012.
\newblock \href {https://aclanthology.org/W12-2103} {Detecting hate speech on
  the world wide web}.
\newblock In \emph{Proceedings of the Second Workshop on Language in Social
  Media}, pages 19--26, Montr{\'e}al, Canada. Association for Computational
  Linguistics.

\bibitem[{Xu et~al.(2021)Xu, Pathak, Wallace, Gururangan, Sap, and
  Klein}]{xu-etal-2021-detoxifying}
Albert Xu, Eshaan Pathak, Eric Wallace, Suchin Gururangan, Maarten Sap, and Dan
  Klein. 2021.
\newblock \href {https://doi.org/10.18653/v1/2021.naacl-main.190} {Detoxifying
  language models risks marginalizing minority voices}.
\newblock In \emph{Proceedings of the 2021 Conference of the North American
  Chapter of the Association for Computational Linguistics: Human Language
  Technologies}, pages 2390--2397, Online. Association for Computational
  Linguistics.

\bibitem[{Zannettou et~al.(2020)Zannettou, Finkelstein, Bradlyn, and
  Blackburn}]{zannettou}
Savvas Zannettou, Joel Finkelstein, Barry Bradlyn, and Jeremy Blackburn. 2020.
\newblock \href {https://ojs.aaai.org/index.php/ICWSM/article/view/7343} {A
  {Quantitative} {Approach} to {Understanding} {Online} {Antisemitism}}.
\newblock \emph{Proceedings of the International AAAI Conference on Web and
  Social Media}, 14:786--797.

\end{thebibliography}
\bibliographystyle{acl_natbib}

\appendix

\section{Appendix}
\label{sec:appendix}
\begin{table*}[!htbp]
\small
    \centering
    \begin{tabular}{c|c|c|c|c|c|c|c}
     \textbf{Model}	&	\textbf{Seed}	&	\textbf{F1}	&	\textbf{Accuracy}	&	\textbf{Balanced Accuracy}	&	\textbf{Recall}	&	\textbf{Precision}	&	\textbf{AUCROC}	\\\hline
Base	&	42	&	0.327	&	0.920	&	0.609	&	0.237	&	0.529	&	0.803	\\
Base	&	44	&	0.389	&	0.929	&	0.632	&	0.276	&	0.656	&	0.834	\\
Base	&	46	&	0.309	&	0.928	&	0.595	&	0.197	&	0.714	&	0.896	\\
Base	&	48	&	0.324	&	0.923	&	0.605	&	0.224	&	0.586	&	0.817	\\
Base	&	50	&	0.257	&	0.919	&	0.578	&	0.171	&	0.520	&	0.844	\\
Base	&	Mean	&	0.321	&	0.924	&	0.604	&	0.221	&	0.601	&	0.839	\\										
KG	&	42	&	0.393	&	0.923	&	0.641	&	0.303	&	0.561	&	0.785	\\
KG	&	44	&	0.355	&	0.925	&	0.618	&	0.250	&	0.613	&	0.827	\\
KG	&	46	&	0.351	&	0.932	&	0.609	&	0.224	&	0.810	&	0.893	\\
KG	&	48	&	0.346	&	0.927	&	0.613	&	0.237	&	0.643	&	0.823	\\
KG	&	50	&	0.237	&	0.923	&	0.569	&	0.145	&	0.647	&	0.832	\\
KG	&	Mean	&	0.336	&	0.926	&	0.610	&	0.232	&	0.655	&	0.832	\\										
    \end{tabular}
    \caption{Results on antisemitic hate speech classification task for Logistic Regression base model (``Base'') vs. KnowledJe enhanced Logistic Regression model (``KG'').}
    \label{tab:lr-results}
\end{table*}

\begin{table*}[!htbp]
\small
    \centering
    \begin{tabular}{c|c|c|c|c|c|c|c}
    \textbf{Model}	&	\textbf{Seed}	&	\textbf{F1}	&	\textbf{Accuracy}	&	\textbf{Balanced Accuracy}	&	\textbf{Recall}	&	\textbf{Precision}	&	\textbf{AUCROC}	\\\hline
Base	&	0	&	0.711	&	0.958	&	0.809	&	0.632	&	0.814	&	0.932	\\
Base	&	1	&	0.629	&	0.950	&	0.751	&	0.513	&	0.813	&	0.909	\\
Base	&	2	&	0.676	&	0.951	&	0.800	&	0.618	&	0.746	&	0.943	\\
Base	&	3	&	0.683	&	0.957	&	0.779	&	0.566	&	0.860	&	0.935	\\
Base	&	4	&	0.699	&	0.954	&	0.819	&	0.658	&	0.746	&	0.924	\\
Base	&	Mean	&	0.680	&	0.954	&	0.792	&	0.597	&	0.796	&	0.929	\\										
KG	&	0	&	0.657	&	0.948	&	0.792	&	0.605	&	0.719	&	0.951	\\
KG	&	1	&	0.631	&	0.948	&	0.762	&	0.539	&	0.759	&	0.927	\\
KG	&	2	&	0.605	&	0.949	&	0.733	&	0.474	&	0.837	&	0.895	\\
KG	&	3	&	0.731	&	0.961	&	0.817	&	0.645	&	0.845	&	0.945	\\
KG	&	4	&	0.662	&	0.949	&	0.793	&	0.605	&	0.730	&	0.903	\\
KG	&	Mean	&	0.657	&	0.951	&	0.779	&	0.574	&	0.778	&	0.924	\\										
    \end{tabular}
    \caption{Results on antisemitic hate speech classification task for DistilBERT base model (``Base'') vs. KnowledJe enhanced DistilBERT model (``KG'')}
    \label{tab:distilbert-results}
\end{table*}

\end{document}